\documentclass[10pt,twocolumn,letterpaper]{article}

\usepackage[pagenumbers]{cvpr} 

\usepackage{graphicx}
\usepackage{amsmath}
\usepackage{amssymb}
\usepackage{booktabs}
\usepackage{makecell}
\usepackage{multirow}
\usepackage{xcolor}
\usepackage{algorithm}
\usepackage{algpseudocode}

\usepackage[pagebackref,breaklinks,colorlinks]{hyperref}

\usepackage[capitalize]{cleveref}
\crefname{section}{Sec.}{Secs.}
\Crefname{section}{Section}{Sections}
\Crefname{table}{Table}{Tables}
\crefname{table}{Tab.}{Tabs.}

\def\cvprPaperID{11683} 
\def\confName{CVPR}
\def\confYear{2022}

\begin{document} 

\title{RAF: Recursive Adversarial Attacks on Face Recognition Using Extremely Limited Queries}

\author{Keshav Kasichainula\\
University Of Houston\\
Houston, USA\\
{\tt\small kkasichainula@uh.edu}
\and
Hadi Mansourifar\\
University Of Houston\\
Houston, USA\\
{\tt\small hmansourifar@uh.edu}
\and
Weidong Shi\\
University Of Houston\\
Houston, USA\\
{\tt\small wshi3@uh.edu}
}
\maketitle

\begin{abstract}
Recent successful adversarial attacks on face recognition show that, despite the remarkable progress of face recognition models, they are still far behind the human intelligence for perception and recognition. It reveals the vulnerability of deep convolutional neural networks (CNNs) as state-of-the-art building block for face recognition models against adversarial examples, which can cause certain consequences for secure systems. Gradient-based adversarial attacks are widely studied before and proved to be successful against face recognition models. However, finding the optimized perturbation per each face needs to submitting the significant number of queries to the target model. In this paper, we propose recursive adversarial attack on face recognition using automatic face warping which needs extremely limited number of queries to fool the target model.  Instead of a random face warping procedure, the warping functions are applied on specific detected regions of face like eyebrows, nose, lips, etc. We evaluate the robustness of proposed method in the decision-based black-box attack setting, where the attackers have no access to the model parameters and gradients, but hard-label predictions and confidence scores are provided by the target model.
\end{abstract}

\begin{table*}[]
\small
\centering
\caption{Classification of previous works on adversarial attacks on face recognition systems. W stands for White Box attack and B stands for Black Box attacks.}
\begin{tabular}{|c|c|l|l|}
\hline
\textbf{Research} & \textbf{Setting} & \multicolumn{1}{c|}{\textbf{Limitation}} & \multicolumn{1}{c|}{\textbf{Feature}} \\ \hline
\cite{sharif2016accessorize} & W & Limited to Eyeglasses & Physically realizable and inconspicuous \\ \hline
\cite{dong2019efficient} & B & Needs 10K queries & Minimum required perturbation \\ \hline
\cite{zhou2018invisible} & W & Limited to infrared perturbations & Invisible to human eyes \\ \hline
\cite{zhu2019generating} & W & Limited to female faces & Realistic make up as perturbation \\ \hline
\cite{nguyen2020adversarial} & W/ B & Limited to light projection & Transformation-invariant pattern generation \\ \hline
\cite{xiao2021improving} & B & Extra regularization techniques are required & Face-like features as adversarial perturbations \\ \hline
\cite{goswami2018unravelling} & B & Limited to unsophisticated adversarial instances & Mimicking real world distortions \\ \hline
\cite{wang2020amora} & B & Needs huge number of queries & Adversarial Morphing Attack \\ \hline
\cite{dabouei2019fast} & B & Low quality adversarial instances & Fast Geometrically-Perturbed Faces \\ \hline
\end{tabular}
\label{tab:related_work}
\end{table*}

\section{Introduction}
Face recognition \cite{liu2017sphereface,deng2019arcface,guo2020learning} is one of the most well-known computer vision tasks which is substantially improved by deep CNNs. Face recognition includes two different tasks : face verification and face identification. Face verification compares a pair of face images to investigate the same identity, while the face identification is a classification task to assign a label as an identity to a face image. Face embeddings are sate of the art which are widely used for both tasks are deep face features which make minimum intra-class and maximum inter-class variances among face classes. Remarkable performance of face embeddings has made face recognition very popular for identity authentication in wide range of security-sensitive applications from financial sectors to criminal identification.
\begin{figure}[H]
\centering
  \includegraphics[width=50mm]{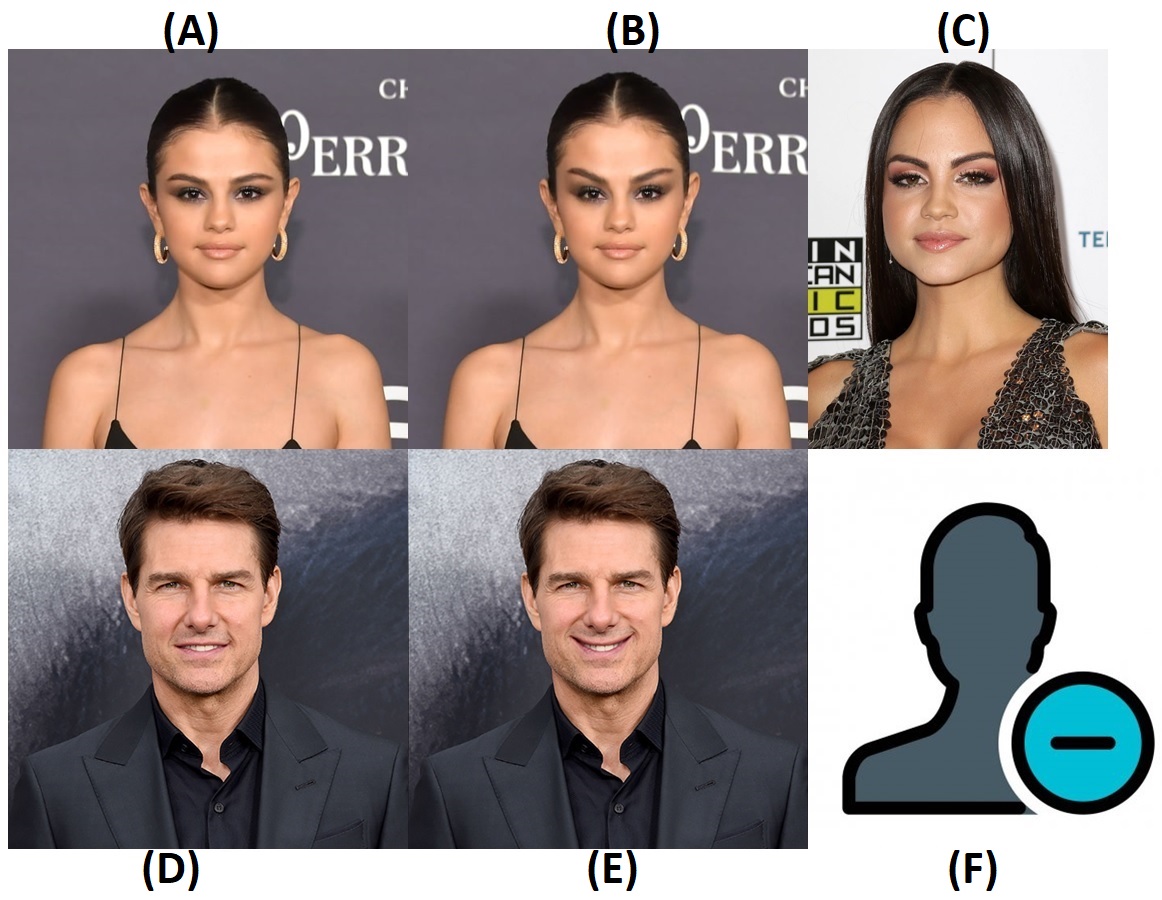}
  \caption{Adversarial dodge and impersonation attacks via automatic face warping : Parts (A,D) show original images, Part(B) is warped image by raise eyebrow function with scale=0.2, Part (C) is the wrong celebrity recognized by AWS Rekognition. Part(E) is warped image using smile function with scale=0.2 and Part(F) demonstrates that no celebrity has been detected by AWS Rekognition. }
  \label{fig:ex_adver_dodger_imper}
\end{figure}

However, deep CNNs proved to be vulnerable against adversarial instances \cite{radiya2021data,fredrikson2015model,mansourifar2020vulnerability,khosravy2021model}. Adversarial inputs \cite{guo2020meets} are defined as maliciously generated indistinguishable instances for human eyes by adding small perturbations\cite{goodfellow2014explaining}. The adversarial attacks against face recognition are often divided to two types: dodging attack which enables the attackers to evade being recognized and impersonation attack in which the attacker is recognized as another individual. For example, one traditional way of adding perturbations to the face is to wear glasses which can fool the target model for null or wrong recognition. To evaluate the threats, the adversarial attacks are categorized under two different settings. In white-box setting, the attackers know the architecture and hyper-parameters of the target model so the loss function can be directly optimized by gradient-based methods. Needless to say that, this scenario is not compatible with real-world cases since the attackers cannot get access to the model details. Black-box setting is more realistic since  no internal model information is known to the attackers except the output of the model including hard-label prediction and confidence score. Designing such adversarial attacks based on perturbation proved to be successful but with a main problem: finding the optimal perturbations needs large number of queries. In recent years, geometrically-perturbed face images proved to be successful to fool the FR \cite{dabouei2019fast,wang2020amora}. However, these approaches are facing a trade-off: the adversarial face images are either low quality or huge number of queries are needed to reach a high quality adversarial instance.  In this paper, we propose Recursive Adversarial Attack on FR   (RAF) using smart face warping to achieve acceptable qualities with extremely limited number of queries. First, we use smart face warping to decompose the face regions into areas with minimum overlap which helps to reduce the size of state space significantly. Second, we form a Depth First Search (DFS) tree with a different warping function as its nodes to search the state space. As a result, a different warping function is applied on the face image as we traverse the DFS tree recursively.  Third, the adversarial instance is submitted to the target model to investigate if the dodging or impersonating goal has been reached given the returned label from the model. The proposed attack has three main characteristics: (i) The adversarial instances are not generated by adding perturbations. (ii) The attacker can reach acceptable dodging or impersonating instances with extremely limited attempts (less than 7 queries). (iii) The adversarial threat can not be removed by pre-processing. To demonstrate the significant success of our proposed attack we use Amazon Rekognition \footnote{https://aws.amazon.com/rekognition/} and different VGGFace models. Amazon Rekognition  is a cloud-based software as a service (SaaS) computer vision platform that was launched in 2016. The technology has been used by law enforcement agencies and was reportedly pitched to Immigration and Customs Enforcement (ICE) in the U.S \cite{bworld1}. Figure \ref{fig:ex_adver_dodger_imper} shows the returned results from Rekognition.
Our contributions are as follows.
\begin{itemize}
\item We propose the first adversarial attack on face recognition using smart face warping.
\item We show that, the state of the art cloud-based paid face recognition service is significantly vulnerable against proposed attack.
\item Our experiments show that, the attacker can reach the dodging or impersonation instances after very limited queries.
\end{itemize}
The rest of paper is organized as follows. Section 2 reviews the related works. Section 3 demonstrates the smart face warping. Section 4 presents the proposed attack. Section 5 provides the experimental results and finally section 6 concludes the paper.

\section{Related Work}
In this section, we review some of the well-known adversarial attacks against face recognition models (FR) which can be categorized to three different types: \textbf {(i)} Adversarial instances are created based on simple noise or real world distortions.
\textbf {(ii)} Adversarial instances are created based on external resources like light projection or infrared.
\textbf{(iii)} Adversarial instances are created based on focusing on face-like features for adversarial perturbations. Despite the simplicity of adding real world distortion or optimized perturbation, they suffer two main problems : they often need large number of queries and they can easily be removed by pre-processing. On the other hand, relying on external light resources to impact the ability of FR models for accurate recognition limits the feasibility of such attacks. Our investigations show that, focusing on face-like features for adversarial perturbations has more capabilities for the next generation of adversarial attacks on FR models. Table \ref{tab:related_work} summarizes the main previous adversarial attacks against FR models. 

\section{Automatic Face Warping}
In this section, we demonstrate the details about the techniques used in the proposed adversarial attack on face recognition. Note that, we used the pychubby\footnote{https://pychubby.readthedocs.io/en/latest/} package for image warping.
\subsection{Image Warping}
Image warping \cite{glasbey1998review,renders2021adjoint} plays a significant role in many applications of image analysis from pre-processing to image augmentation. A warping is a pair of two-dimensional functions, $u(x,y)$ and $v(x,y)$, which map a position $(x,y)$
from one image to position $(u,v)$ in another image, where $x$ denotes column number and $y$ denotes row number. In the other word, Given source image $I$ and the correspondence between the original position $p_i = (x_i, y_i)^T$ of a point in $I$ and its desired
new position $qi = (u_i, v_i)^T$, where $i = 1, . . . , n$, the warping function generates an image $I'$ such that $I'(q_i) = I(p_i)$ for each point $i$, where, $I_i$ and $I'_i$
represent the intensities or colors of the images. In case of face warping, there have been many approaches to find an appropriate warp. Relying on facial landmarks is a practiced method and it can provide a robust road map for automatic face warping. Our observations show that, even a negligible amount of transformation of certain landmarks as base warping rule can create adversarial instances \cite{dong2018boosting} which are malicious enough for dodge and impersonation attacks. 
\begin{figure}[H]
\centering
  \includegraphics[width=40mm]{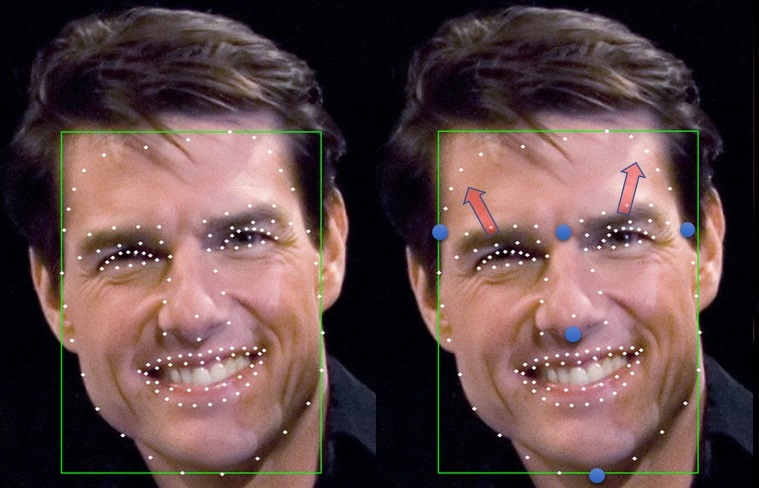}
  \caption{68 facial landmarks and five reference points for affine transformation. The arrows illustrate land mark transformation for Raise Eyebrow warping function.}
  \label{fig:landmark}
\end{figure}
\subsection{Face Landmark Detection}
Let $I$ be an input image of size $W \times H \times C $, where $W$ is width, $H$ is height and $C$ is number 
of image color channels. Then facial landmark detection problem \cite{khabarlak2021fast} is to find a function $\Phi : I\rightarrow L$, which from the input image $I$ predicts a landmark vector $L$ where each landmark contains $x$ and $y$ coordinates corresponding to a specific predefined facial point. Number of landmarks can be 
different depending on the method. In this paper, we used \cite{kazemi2014one} which utilizes a cascade of regressors to extract 68 facial landmarks. 
\subsection{Automatic Face Warping}
After detecting 68 face landmarks, a set of predefined transformations are applied per each warping functions. To update the location of neighboring face landmarks, five selected landmarks are used to estimate an affine transformation \cite{dixit2021fast} between the reference and input space. This transformation is encoded in a matrix to transform the points from reference to target space and vice versa via a simple matrix multiplication.
Figure \ref{fig:landmark} (left) shows the 68 landmarks and the right image shows 5 selected landmarks illustrated by blue large dots which used for affine transformation. 

\begin{figure}[]
\small
\centering
    \includegraphics[width=80mm]{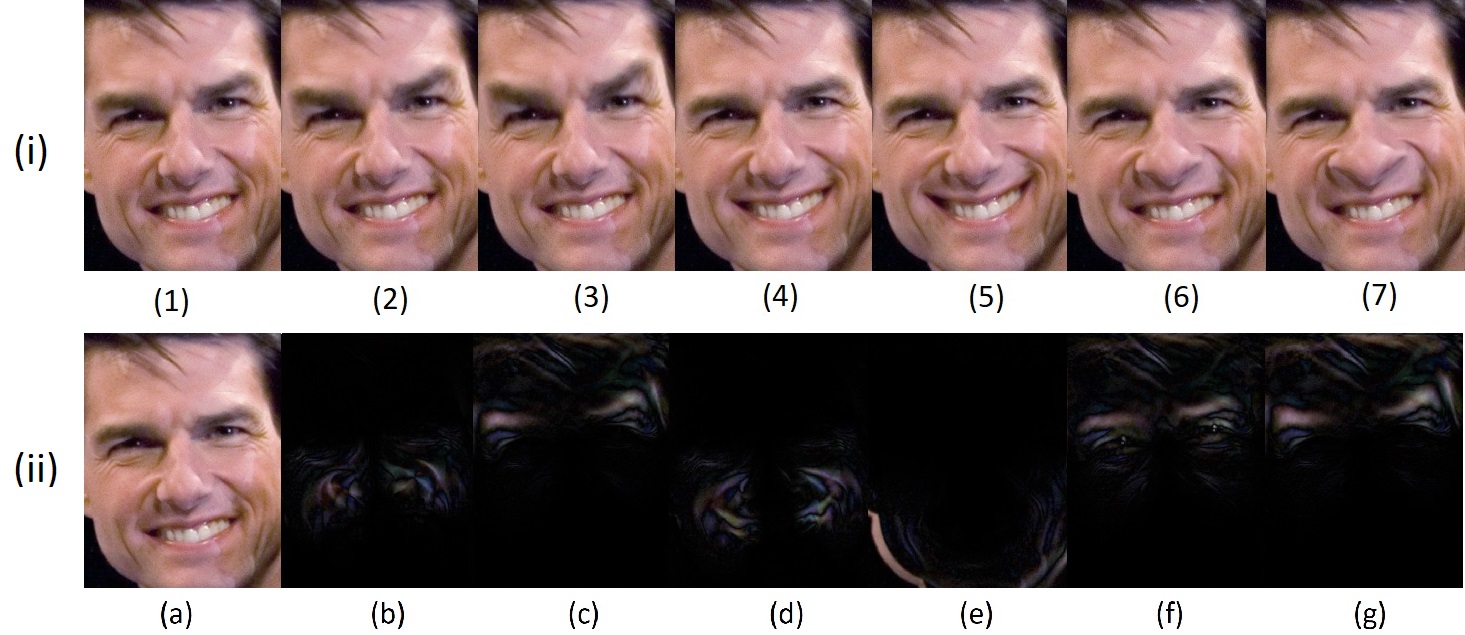}
  \caption{(i) A set of adversarial instances generated by proposed method. (ii) The impacted facial regions by automatic warping functions with respect to original image.}
  \label{fig:adver_instance_impact_regions}
\end{figure}

\begin{table}
\small
\centering
\caption{Warping functions used to generate images in Figure 3(i) along with scale and recognition status by AWS Rekognition. RE denotes Raised Eyebrows and SN denotes Stretched Nose}
\label{tab:func_scale}
\begin{tabular}{|l|c|c|c|c|c|c|c|} 
\hline
\textbf{Number}     & 1            & 2            & 3            & 4              & 5              & 6            & 7             \\ 
\hline
\textbf{Function}   & \textit{RE}  & \textit{RE}  & \textit{RE}  & \textit{Smile} & \textit{Smile} & \textit{SN}  & \textit{SN}   \\ 
\hline
\textbf{Scale}      & \textit{0.1} & \textit{0.2} & \textit{0.3} & \textit{0.1}   & \textit{0.2}   & \textit{0.1} & \textit{0.2}  \\ 
\hline
\textbf{Recognized} & \textit{Yes} & \textit{No}  & \textit{No}  & \textit{Yes}   & \textit{No}    & \textit{No}  & \textit{No}   \\
\hline
\end{tabular}
\end{table}

\subsection{Warping Functions}
Finding an optimized adversarial face by targeting all the face regions has two main problems: (i) It takes a huge number of queries to reach an adversarial face. (ii) The quality of adversarial face is not guarantied. However, automatic face warping functions can avoid these problems since (i) Each warping function targets a specific face region. (ii) Each warping function can reach the adversarial instances just by targeting its own target area with extremely limited queries. (iii) The quality of each warping function can be monitored (iv) Setting a boundary as warping scale is easy for each warping function to gauarantee the warped faces quality. We used following automatic warping function:
\begin{itemize}
\item Raise Eyebrow : This function mostly impacts the region between eyebrow and upper forehead. Parts (c,g) of Figure \ref{fig:adver_instance_impact_regions}(i) shows the changes applied by raise eyebrow function with scales 0.3 and 0.4, respectively comparing to original face shown in part (a) of Figure \ref{fig:adver_instance_impact_regions}(ii). Also, right image of Figure \ref{fig:landmark} illustrates the two landmarkes used for raised eyebrow function.
\item Smile : This function mostly impacts the upper corner lips and mouth regions as shown in part (b) of figure \ref{fig:adver_instance_impact_regions}(ii).
\item Nose Stretch : This function mostly impacts the regions surrounding the nose as shown in part (d) of figure \ref{fig:adver_instance_impact_regions}(ii).
\item Chubbify : This function impacts the most distant face landmarks to the nose as shown in part (e) of figure \ref{fig:adver_instance_impact_regions}(ii).
\item Open Eyes : This function mostly impacts the surrounding regions to the eyes and eyebrows as shown in part (f) of figure \ref{fig:adver_instance_impact_regions}(ii).
\end{itemize}
Table \ref{tab:func_scale} shows the warping functions, the scale used and the results from the AWS Rekogniton for in the figure \ref{fig:adver_instance_impact_regions}(i).
\subsection{Affine Transformation}
\textit{Definition.} $T : R\rightarrow R^n$ is an affine transformation \cite{sayed2021trajectory} if $T$ is a homogeneous linear transformation followed by a translation. Namely, there a matrix $A$ and vector \textbf{b} such that, $T$(\textbf{x})=\textbf{x}$A$ + \textbf{b} for all \textbf{x}. In the other word, an affine transformation is any transformation that preserves collinearity and can be combination of linear transformations and translations.
Properties of affine transformations are as follows. Origin does not necessarily map to origin, lines map to lines, parallel lines remain parallel and ratios are preserved. To preserve the lines in both images, a regression algorithm is used to estimate the linear parameters which minimised the sum of squared differences between the spots:
\begin{equation}
\sum_{i=1}^{m}\begin{Bmatrix}
u_i -u(x_i,y_i)
\end{Bmatrix} + \sum_{i=1}^{m}\begin{Bmatrix}
v_i -v(x_i,y_i)
\end{Bmatrix} 
\end{equation}
, where $( x_1 , y_1 ),..., (x_m,y_m )$ are in the  first image, and $( u_1,v_1 ),...,(u_m,v_m )$
are in the warped image \cite{glasbey1998review}.

\section{Adversarial Face Warping Attack}
In this section, we present the proposed
black-box attack setting and evolutionary attack method against a face recognition model.
\subsection{Attack Setting}
A target face recognition model is denoted by $f(x):\chi \rightarrow \gamma \; (\chi \subset \mathbb{R}^n)$, where in face identification task the input image $x$ is compared
with a gallery set of face images, and then classifier assigns the $x$ a specific identity such that, $\gamma = {1,2,...,K}$ where $K$ is the number of identities. Supposing a target face image $x$ and correct label $\gamma$, the goal of attacker is to warp $x$ to create an adversarial face image $x^w$ and get label $\gamma^w$ from model such that, $\gamma^w \neq \gamma$ with minimum changes
comparing to $x$. It can be obtained by solving a constrained optimization problem as follows. 
\begin{equation}
\underset{x^w}{min} \; \Delta (x,x^w), \; s.t.\, \gamma \neq \gamma^w 
\end{equation}
We use the $L_2$ distance to calculate the $\Delta$. The constrained problem in Equation (2) can be equivalently reformulated as the following unconstrained optimization problem.
\begin{equation}
\underset{x^w}{min} \,  \mathcal{L} (x^w) =\: \Delta(x,x^w) + \delta (\gamma \neq \gamma^w ) 
\end{equation}
where $\delta(x,x^w) = 0$ if $\gamma \neq \gamma^w$, otherwise $\delta(x,x^w) = \infty $ \\
As a result, the adversarial instance $x^w$ is supposed to be obtained with minimum required perturbation by optimizing Eq. (3). 
\subsubsection{Dodging}
 This attack corresponds to generating an adversarial image that is not recognized as any of the identities in the training data. 
 \begin{equation}
\exists x_{i}^{w} \;, \; \; s.t.\;  \forall \gamma  (\gamma_i \neq \gamma ^w)
\end{equation}

\subsubsection{Impersonation} This attack corresponds to generating an adversarial image that is recognized as a wrong identity in the training data. 
 \begin{equation}
\forall x_i \exists x_{j}^w (\gamma_i = \gamma_{j} ^w), \; \; s.t.\; i \neq j
\end{equation}
\begin{figure*}[]
\centering
  \includegraphics[width=120mm]{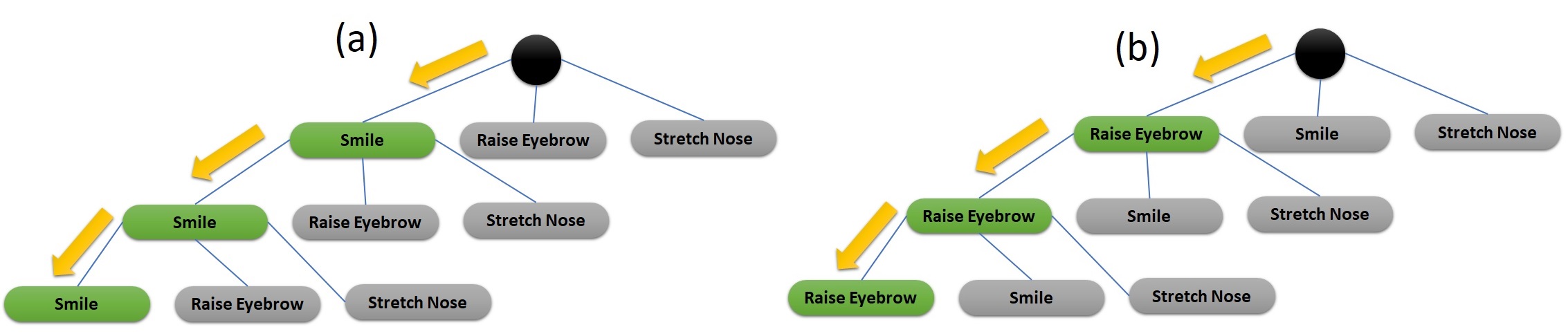}
  \caption{An illustration of DFS tree with different node orders: The order of warping functions can impact the search results. }
  \label{fig:dfs}
\end{figure*}
\subsection{Recursive Adversarial Attack}
One of the most significant deficiencies of old adversarial attacks is that, the output is insensitive to small input perturbations. That's why the attacker needs huge number of queries to reach a successful adversarial instance and the gradient estimation methods cannot be directly used.
Some methods \cite{maheshwary2021generating,saadatpanah2020adversarial} successfully reformulated the discontinuous optimization problem in Eq. (2) as some continuous
optimization problems and use gradient estimation methods for optimization. But they need to calculate the distance of a point to the decision boundary or estimate the predicted probability by the hard-label outputs, which are less efficient. In this paper, we use recursive optimization to find the minimum warping required to fool the target model. The recursive optimization is an evolutionary
computation framework that can be used to solve high dimensional optimization problems via a ‘divide-and-conquer’ mechanism, where the main challenge lies in problem decomposition.
\subsubsection{Definition1}
$f(x)$ is globally decomposable if there exists a partition $\begin{Bmatrix} x_C,& x_{U_1}, & x_{U_2}
\end{Bmatrix}$  such that, for every partial assignment $\rho c, f|_{\rho c}(x_{U_1},x_{U_2})= f_1|_{\rho c}(x_{U_1})+ f_2|_{\rho c}(x_{U_2}) $.
According to Definition 1 \cite{friesen2015recursive}, face warping problem can be decomposed to a set of sub-functions like smile, raise eyebrow, nose stretch,etc, since each of the the sub-functions are independent from each other with minimum overlap in terms of impacted face regions. As a result, (i) face warping functions enable the attacker to avoid the random perturbations by applying smart face warping. (ii) The state space can be decomposed to non-overlapping regions by applying sub-function recursively.
\begin{figure}[H]
\centering
  \includegraphics[width=75mm]{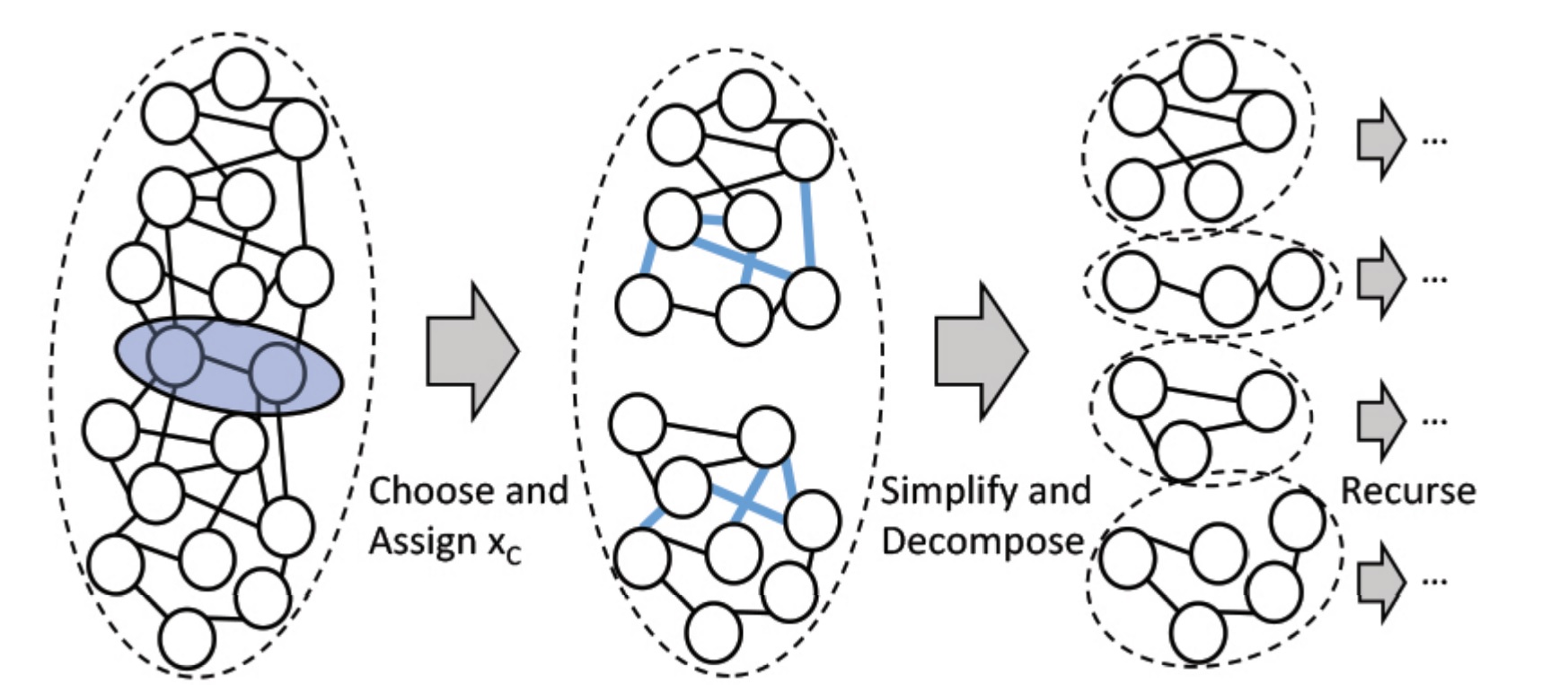}
  \caption{Decomposing the warping regions into the independent regions \cite{friesen2015recursive}.}
  \label{fig:decomposing}
\end{figure}
Figure \ref{fig:decomposing} illustrates the decomposition of the state space into smaller regions by divide and conquer strategy \cite{friesen2015recursive}. Each node represents a different warping region to be processed by a different warping function with different operation and scale. 


        
            
  
\begin{algorithm}
\caption{RAF: Recursive Adversarial Attack using Face Warping and Limited Queries}
\label{alg:dfs}
\begin{algorithmic}
    \Function{REFA}{$x$, $WF_i$,$scale$,$step$,$id$,$l$}
            \State $x = WF_i(x,scale)$
            \State $rid = submit(x)$
            \If{(id==rid) and (i$>$l)} 
                \State print(No Adversarial Instance Found)
                \State \textbf{return} x
            \ElsIf{(id!=rid)}
                \State print(Adversarial Instance Found)
                \State \textbf{return} x
            \EndIf
            \State $i = i + 1$
            \State \textbf{return} REFA(x,$WF_i$,scale,step,id)
    \EndFunction
\end{algorithmic}
\end{algorithm}

\subsection{Recursive Attack Analysis}
In this section we demonstrate the superiority of proposed method, Recursive Adversarial Attack on FR (RAF) versus standard warping algorithms to create adversarial instances. Formally, we investigate the required state space to be explored by RAF to show that it needs exponentially less time than the same optimizer which can converge to the global optimum
based on non-smart warping functions.
\subsubsection{Complexity}
Like RDIS algorithm \cite{friesen2015recursive} RAF begins by choosing \(x_c\) as a block of variables. Assuming
that this choice is made heuristically using the PaToH library \cite{ccatalyurek2011patoh}
for hypergraph partitioning, which is a multi-level technique,
then the complexity of choosing variables is linear\cite{nocedal2006numerical}. Within the loop, RAF chooses values
for \(x_C\) , simplifies and decomposes the function, and finally
recurses. Let the complexity of choosing values using the
subspace optimizer be \(g(d)\), where  \(|x_c|= d\), and let one call to the subspace optimizer be cheap relative to $n$ (e.g. computing the gradient of $f$ with respect to \(x_c\) or taking a step on a grid). Simplification requires iterating through the set of terms and computing bounds, so it is linear in the number
of terms,  \(m\). The connected components are maintained by
a dynamic graph algorithm \cite{holm2001poly} which has an
amortized complexity of  \(O{(\log2}(|V |))\) per operation, where
 \(|V |\) is the number of vertices in the graph. Finally, let the
number of iterations of the loop be a function of the dimension, \(\xi(d)\), since more dimensions generally require more
restarts.
\subsubsection{Proposition1.} If at each level, RAF chooses $x_C \subseteq  x$ of size
$|x_C |= d$ such that, for each selected value $\rho_C$ , the simplified function $f|\rho_C$
$(x_U )$ locally decomposes into $k > 1$ independent sub-function $ \begin{Bmatrix}
f_i (x_{U_i})
\end{Bmatrix}$ with equal-sized domains $x_{U_i}$
, then the time complexity of RAF is $O( \frac{n}{d}\xi (d)^{log_k \frac{n}{d} })$ \cite{friesen2015recursive} with following proof \cite{bworld2}

Proof. Assuming that is of the same order as \(n\),
the recurrence relation for RDIS is 
\begin{equation} \label{eq11}
\begin{split}
T(n) & = O(n) +\xi(d)[g(d) + O(m) + O(n)  \\
 & +O(d {\log^2 (n)}) + kT(\frac{n-d}{k})]
\end{split}
\end {equation}
which can be simplified to \[T(n) = \xi(d)[KT(\frac{n}{k})+ O(n)]+O(n)\]
Noting that the recursion halts at \(T(d)\), the solution to the above recurrence relation is then 
\begin{equation} \label{eq1}
T(n)=c_1(k\xi (d))^{log_k(\frac{n}{d})}
+c_2 n \sum_{r=0}^{\log_k(\frac{n}{d})-1}\xi(d)^r 
\end {equation}
which is 

\[O(k\xi (d))^{log_k(\frac{n}{d})}= O (\frac{n}{d}\xi(d)^{\log_k(\frac{n}{d})})\]
Algorithm \ref{alg:dfs} shows the required steps for recursive adversarial attack on face recognition, where $WF_i$ represents the active warping function which applies the automatic warping on given image denoted by $x$. First, we create a warping function set containing a sequence of warping functions. What we need to do is to traverse a state space of warping functions and apply them one by one till finding an adversarial instance. Since we want to find adversarial instance in extremely limited queries, we set a boundary denoted by $l$. Also the $id$ is used to check the original target identity with the returned identity from target model ($rid$). The recursive algorithm stops in case $id \neq rid$ which means a successful impersonation attack but if $rid = Null$ then it's a successful dodge attack. The other condition to stop is number of queries.


\begin{table*}[]
\scriptsize
\centering{}
\caption{Comparing different target FR models using RAF. The numbers bellow indicate the number of dodge and impersonating results received from all the warped images submitted to the models in less than 4 queries. RE - Raised Eyebrows and SN - Stretched Nose.}
\label{tab:succ_attack_less_4_queries}
\begin{tabular}{c|ccccccccc|ccccccccc|}
\cline{2-19}
\multirow{3}{*}{} & \multicolumn{9}{c|}{\textbf{Dodge Attack}} & \multicolumn{9}{c|}{\textbf{Impersonation Attack}} \\ \cline{2-19} 
 & \multicolumn{3}{c|}{\textit{CelebA}} & \multicolumn{3}{c|}{\textit{Web Casia}} & \multicolumn{3}{c|}{\textit{Collected}} & \multicolumn{3}{c|}{\textit{CelebA}} & \multicolumn{3}{c|}{\textit{Web Casia}} & \multicolumn{3}{c|}{\textit{Collected}} \\ \cline{2-19} 
 & \multicolumn{1}{c|}{Smile} & \multicolumn{1}{c|}{RE} & \multicolumn{1}{c|}{SN} & \multicolumn{1}{c|}{Smile} & \multicolumn{1}{c|}{RE} & \multicolumn{1}{c|}{SN} & \multicolumn{1}{c|}{Smile} & \multicolumn{1}{c|}{RE} & SN & \multicolumn{1}{c|}{Smile} & \multicolumn{1}{c|}{RE} & \multicolumn{1}{c|}{SN} & \multicolumn{1}{c|}{Smile} & \multicolumn{1}{c|}{RE} & \multicolumn{1}{c|}{SN} & \multicolumn{1}{c|}{Smile} & \multicolumn{1}{c|}{RE} & SN \\ \hline
 
\multicolumn{1}{|c|}{\textit{\textbf{AWS Rekognition}}} & \multicolumn{1}{c|}{47} & \multicolumn{1}{c|}{16} & \multicolumn{1}{c|}{184} & \multicolumn{1}{c|}{24} & \multicolumn{1}{c|}{4} & \multicolumn{1}{c|}{147} & \multicolumn{1}{c|}{33} & \multicolumn{1}{c|}{34} & \multicolumn{1}{c|}{68} & \multicolumn{1}{c|}{47} & \multicolumn{1}{c|}{16} & \multicolumn{1}{c|}{3} & \multicolumn{1}{c|}{24} & \multicolumn{1}{c|}{4} & \multicolumn{1}{c|}{3} & \multicolumn{1}{c|}{12} & \multicolumn{1}{c|}{34} & \multicolumn{1}{c|}{26} \\ \hline

\multicolumn{1}{|c|}{\textit{\textbf{VGG 16}}} & \multicolumn{1}{c|}{47} & \multicolumn{1}{c|}{56} & \multicolumn{1}{c|}{48} & \multicolumn{1}{c|}{46} & \multicolumn{1}{c|}{47} & \multicolumn{1}{c|}{23} & \multicolumn{1}{c|}{40} & \multicolumn{1}{c|}{34} & \multicolumn{1}{c|}{35} & \multicolumn{1}{c|}{36} & \multicolumn{1}{c|}{37} & \multicolumn{1}{c|}{16} & \multicolumn{1}{c|}{60} & \multicolumn{1}{c|}{38} & \multicolumn{1}{c|}{23} & \multicolumn{1}{c|}{21} & \multicolumn{1}{c|}{17} &  \multicolumn{1}{c|}{9}\\ \hline

\multicolumn{1}{|c|}{\textit{\textbf{Resnet50}}\cite{theckedath2020detecting}} & \multicolumn{1}{c|}{24} & \multicolumn{1}{c|}{10} & \multicolumn{1}{c|}{19} & \multicolumn{1}{c|}{29} & \multicolumn{1}{c|}{34} & \multicolumn{1}{c|}{24} & \multicolumn{1}{c|}{22} & \multicolumn{1}{c|}{24} & \multicolumn{1}{c|}{18} & \multicolumn{1}{c|}{1} & \multicolumn{1}{c|}{8} & \multicolumn{1}{c|}{1} & \multicolumn{1}{c|}{9} & \multicolumn{1}{c|}{11} & \multicolumn{1}{c|}{3} & \multicolumn{1}{c|}{24} & \multicolumn{1}{c|}{19} & \multicolumn{1}{c|}{9} \\ \hline

\multicolumn{1}{|c|}{\textit{\textbf{Senet50}}\cite{luo2019cross}} & \multicolumn{1}{c|}{16} & \multicolumn{1}{c|}{43} & \multicolumn{1}{c|}{40} & \multicolumn{1}{c|}{15} & \multicolumn{1}{c|}{26} & \multicolumn{1}{c|}{21} & \multicolumn{1}{c|}{44} & \multicolumn{1}{c|}{61} & \multicolumn{1}{c|}{53} & \multicolumn{1}{c|}{5} & \multicolumn{1}{c|}{6} & \multicolumn{1}{c|}{4} & \multicolumn{1}{c|}{6} & \multicolumn{1}{c|}{15} & \multicolumn{1}{c|}{8} & \multicolumn{1}{c|}{27} & \multicolumn{1}{c|}{23} & \multicolumn{1}{c|}{14} \\ \hline
\end{tabular}
\end{table*}

\begin{table*}
\scriptsize
\centering
\caption{Detail description of adversarial instances presented in part (A) of Figure \ref{fig:5_6_queries}. RE denotes Raised Eyebrows, SN denotes Stretch Nose, D denotes Dodge Attack, I denotes Impersonating Attack and N denotes no attack.}
\label{tab:detailed_adversarial_instances}
\begin{tabular}{|c|c|c|c|c|c|c|c|c|c|} 
\hline
\multirow{2}{*}{\textit{\textbf{Figure}}} & \textit{\textbf{Image Number}}                  & \textbf{(1)}      & \textbf{(2)}      & \textbf{(3)}      & \textbf{(4)}                                         & \textbf{(5)}                                         & \textbf{(6)}         & \textbf{(7)}                                         & \textbf{(8)}       \\ 
\cline{2-10}
                                          & \textit{\textbf{Function(Scale)}}               & \textit{RE (0.1)} & \textit{RE (0.2)} & \textit{RE (0.3)} & \textit{Smile (0.1)}                                 & \textit{Smile (0.2)}                                 & \textit{Smile (0.3)} & \textit{SN (0.1)}                                    & \textit{SN (0.2)}  \\ 
\hline
\textbf{(a)}                              & \multirow{5}{*}{\textit{\textbf{Attack Type}}~} & \textit{N}        & \textit{N}        & \textit{D}        & \textit{N}                                           & \textit{D}                                           & \textit{D}           & \textit{D}                                           & \textit{I}         \\ 
\cline{1-1}\cline{3-10}
\textbf{(b)}                              &                                                 & \textit{I}        & \textit{D}        & \textit{D}        & \textit{N}                                           & \textit{D}                                           & \textit{D}           & \textit{D}                                           & \textit{I}         \\ 
\cline{1-1}\cline{3-10}
\textbf{(c)}                              &                                                 & \textit{N}        & \textit{N}        & \textit{N}        & \textit{N}                                           & \textit{D}                                           & \textit{I}           & \textit{D}                                           & \textit{D}         \\ 
\cline{1-1}\cline{3-10}
\textbf{(d)}                              &                                                 & \textit{N}        & \textit{N}        & \textit{D}        & \begin{tabular}[c]{@{}c@{}}\textit{N}\\\end{tabular} & \textit{D}                                           & \textit{D}           & \textit{N}                                           & \textit{D}         \\ 
\cline{1-1}\cline{3-10}
\hline
\end{tabular}
\end{table*}

\subsection{DFS Search Tree}
Although the recursive adversarial search is supposed to traverse all the warping functions, the order of nodes in the DFS tree can impact the required number of queries in some faces. That's why we need to investigate which one of the warping functions has more impact on adversarial attacks. Our empirical experiments show that, three warping functions including Smile, Raise Eyebrow and Stretch Nose have the highest impacts with high quality warped faces. Thus, we decided to keep the DFS tree limited to three mentioned functions as shown in Figure \ref{fig:dfs}.

\subsection{Hybrid Adversarial Faces}
\begin{figure}[]
\centering
  \includegraphics[width=50mm]{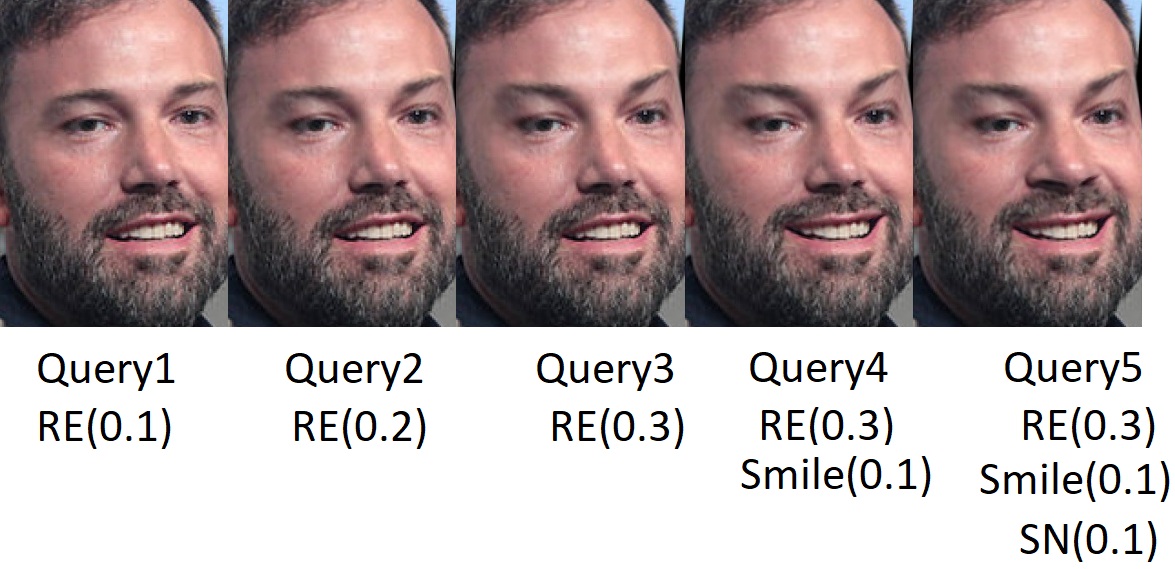}
  \caption{successful dodge attack with 5 queries: A hybrid adversarial instance generated by three different warping functions. }
  \label{fig:Combined_images_less_4}
\end{figure}

Although many adversarial instances can be generated using only one warping functions, sometimes more than one warping functions are needed to fool the target mode. Such face images are called hybrid adversarial instances as shown in Figure \ref{fig:Combined_images_less_4}.
\section{Experiments}
In this section we present the experimental setups, results and discussions related to proposed adversarial attacks on face recognition.
\subsection{Datasets}
For the sake of experiments, we used following datasets.
\textbf {Collected Dataset}: We collected 101 casual celebrity faces from the web which mostly contains frontal celebrity faces.
In addition, we used \textbf{CelebA Dataset} \cite{liu2018large} and \textbf{CASIA-WebFace Dataset} \cite{yi2014learning}.
\subsection{Target Models}
\begin{figure}[H]
\centering
  \includegraphics[width=70mm]{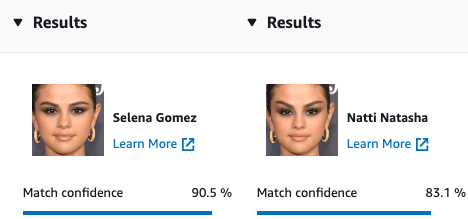}
  \caption{Success evidence: a screenshot of AWS Rekognition demo version output representing a successful impersonation attack (Warping Function=Raised Eyebrow, Scale=0.2).}
  \label{fig:selena_gomez}
\end{figure}

In addition to creating FR models using \textbf{VGG16}, \textbf{Resnet50} \cite{theckedath2020detecting} and \textbf{Senet50} \cite{luo2019cross}, we also used following target models. \textbf{AWS Rekognition} A cloud-based software as a service computer vision platform. Figure \ref{fig:selena_gomez} shows a sample result using AWS Rekognition. \textbf{VGGFace} A series of models developed for FR by Visual Geometry Group (VGG). \textbf{ BetaFace}\footnote{https://www.betafaceapi.com}: It is an online face classifier which provides verification (faces comparison) and identification (faces search) services.

\subsection{Evaluation of Warping Functions Impact}
In this section, we evaluate different smart warping functions to find which one has more success for both dodge and impersonation attacks. As mentioned earlier, the order of nodes as warping functions can change the search path significantly. To find the best possible order of warping functions we need to evaluate the impact of each warping functions. To do so, we tested each function as the first acting function in the DFS tree. For the VGG16, Resnet50, Senet50 we selected thresholds for confidence scores in case of impersonating and dodge attacks. For collected datasets in all models we selected the threshold of to be between 94.99\% to 50\% for dodge and for impersonating it to be above 95\%. In case of CelebA dataset in case of VGG16 and sesnet50 we selected the threshold similar to collected but in case of Resnet50 for impersonation above 70\% and for dodge below 1\%. For Web Caisia in case of VGG16 and Senet 50 the threshold was above 98\% for impersonating and between 97.9\% to 80\% in case of dodge, for Resnet50 it is similar to collected dataset .Table \ref{tab:succ_attack_less_4_queries} summarizes the number of successful attacks per each warping function. We can summarize the evaluation results as follows. \textbf{(i)} The Nose stretch function reaches the best results in case of dodge attack in all three tested datasets. \textbf{(ii)} In case of impersonation attacks, a different warping function shows the superiority on each dataset.\textbf{(iii)} In overall, the most impactful warping functions are sorted as Stretch Nose, Smile and Raise Eyebrow in terms of both attack types and in all tested datasets. The results for 5 and 6 queries could be found in the supplementary materials.

\begin{figure*}[]
\centering
  \includegraphics[width=140mm]{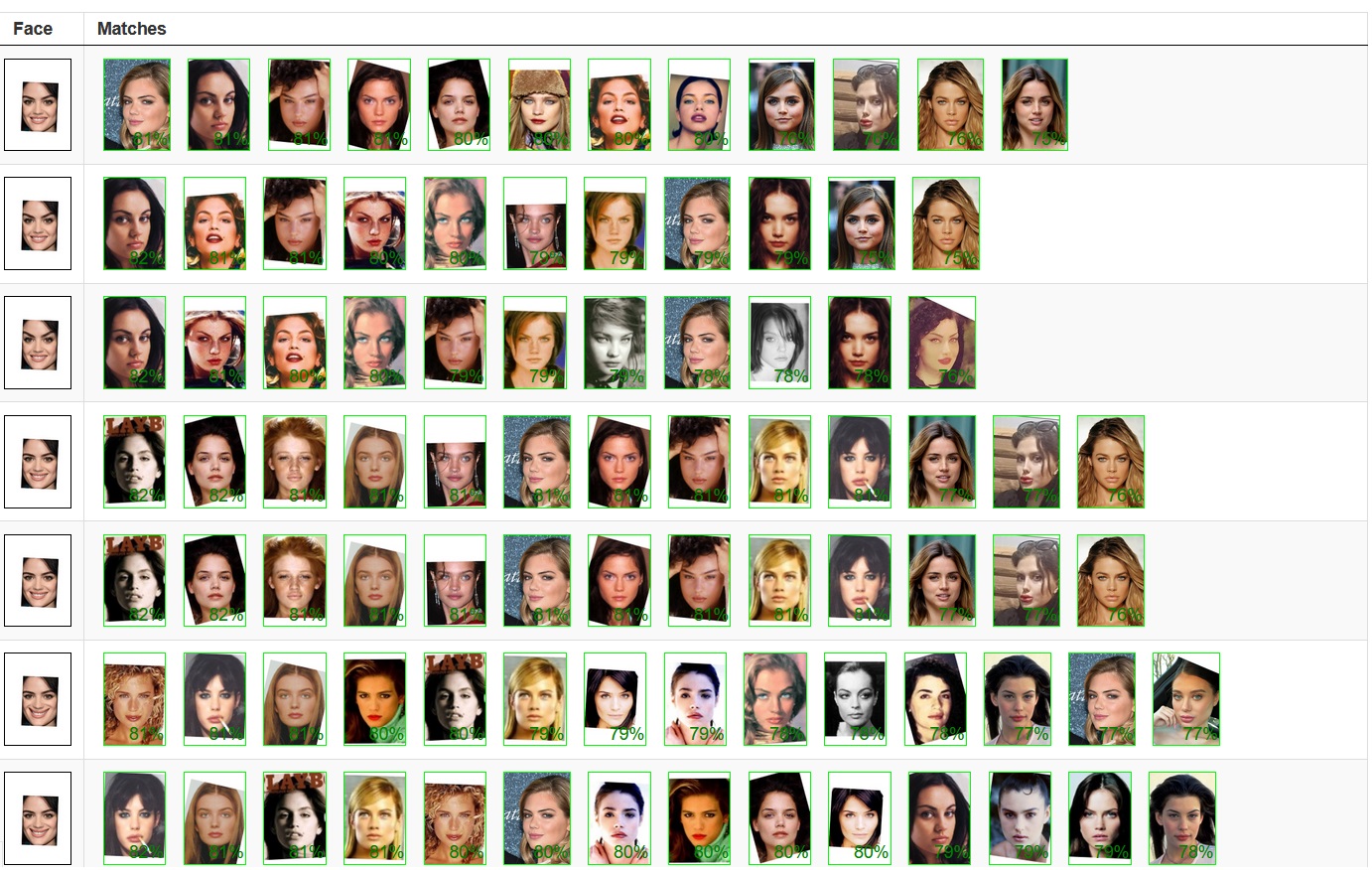}
  \caption{Experimental results on Betaface model: the impact of automatic face warping on a face retrieval model. A minor change in warping can significantly impact the identity and  order of retrieved celebrities. Warped images generated by two different functions including raise eyebrow and smile.}
  \label{fig:openFace}
\end{figure*}

\begin{figure*}[]
\centering
  \includegraphics[width=150mm]{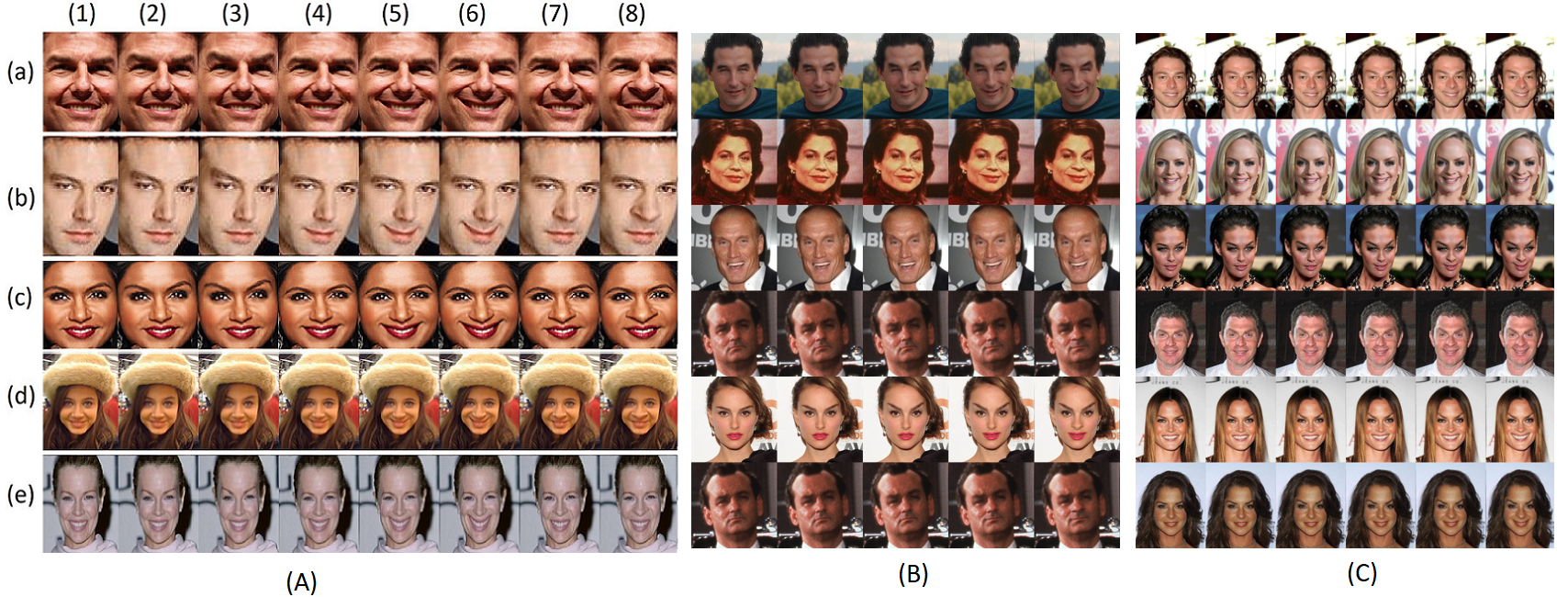}
  \caption{(A) Adversarial instances created only by one warping function. Hybrid attack with (B) 5 queries and (C) 6 queries : The successful adversarial instances found in fifth and sixth queries respectively (the last columns in part (B),(C). }
  \label{fig:5_6_queries}
\end{figure*}

\subsection{Attack Evaluation with Extremely Limited Queries  }
To evaluate the proposed method we selected the most challenging scenario in which the attacker has only three chances to fool the target model. In the other word, we assumed that the target model accepts only three queries from each authorized user in a specific period of time. It means that, the DFS tree looks like the highlighted path by orange arrows in Figure \ref{fig:dfs}. Our experimental results show that, the proposed adversarial attack using the automatic face warping reaches high success rate with extremely limited queries. We can summarize the limited query results as follows. \textbf{(i)} In case of dodge attack in collected dataset, the majority of successful attacks find the adversarial instance with only one query.
\textbf{(ii)} In case of dodge attack in CelebA and CASIA Web datasets, the majority of successful attacks find the adversarial instance with two queries.
\textbf{(iii)} In case of impersonation attack in collected dataset, the majority of successful attacks find the adversarial instance with three queries.
\textbf{(iv)} In case of impersonation attack in CelebA dataset, the majority of successful attacks find the adversarial instance with two queries.
\textbf{(v)} In all tested datasets, finding the impersonation instance is very challenging comparing to dodge instances.

\subsection{Adversarial Instance Samples}
In this section, we present some successful adversarial instances in both dodge and impersonation categories created by our proposed method. Figure \ref{fig:selena_gomez} shows a print screen of AWS Rekognition page with an original face correctly recognized and a wrongly recognized warped face which obtained by two queries (Raise eyebrow, Scale=0.2). Due to page limit we present a limited number of results here but more results are shown in supplementary materials. Also, Part (A) of Figure \ref{fig:5_6_queries} shows some failed and successful adversarial instances and the details can be found in Table \ref{tab:detailed_adversarial_instances}. Note that, parts (a,b,c) are selected from collected dataset while part (d,e) are from CelebA and CASIA Web datasets, respectively.

\subsection{Hybrid Adversarial Instances}
Although, we selected the most challenging scenario to find the adversarial instances with less than 4 queries, we also studied the warped faces which need more than three queries to fool the target model. Such warped faces are called hybrid adversarial instances since more than one warping function contribute to warp them. Figure \ref{fig:Combined_images_less_4} shows a case study which took 5 queries to reach a successful dodge attack. 

\subsection{Experiments on Betaface model}
Betaface enables us to browse the synthesized faces in wide range of faces. Using the image retrieval tool we can keep track of warped faces and their impacts in the retrieved faces comparing to corresponding original faces. As face retrieval is different from face recognition it is important to evaluate the face warping impact on face retrieval tools. Experiment results can be found in Figure \ref{fig:openFace}.
\subsection{Ablation Study}
In this section, we compare the impact of RAF with the closest related works \cite{dabouei2019fast,wang2020amora} in terms of impact on Face retrieval models.

\section{Conclusion}
In this paper, we proposed a new adversarial attack on face recognition models with extremely limited queries. To do so, we used automatic face warping in a recursive adversarial context to target the commercial and public face recognition-retrieval models. Our experiments showed that, the proposed attack can fool the online FR models with less than 4 queries.


{\small
\bibliographystyle{ieee_fullname}
\bibliography{ref}
}

\end{document}